\documentclass[letterpaper]{article} 
\usepackage{aaai23}  
\usepackage{times}  
\usepackage{helvet}  
\usepackage{courier}  
\usepackage[hyphens]{url}  
\usepackage{graphicx} 
\usepackage{natbib}  
\usepackage{caption} 
\usepackage{algorithm}
\usepackage{algorithmic}

\usepackage{newfloat}
\usepackage{listings}
\DeclareCaptionStyle{ruled}{labelfont=normalfont,labelsep=colon,strut=off} 
\floatstyle{ruled}
\newfloat{listing}{tb}{lst}{}
\floatname{listing}{Listing}
\usepackage{microtype}
\usepackage{subfigure} 
\usepackage{latexsym}
\usepackage{booktabs}
\usepackage{multirow}
\usepackage{makecell}
\usepackage{amsmath}
\usepackage{amssymb}
\usepackage{enumerate}

\title{Revising Image-Text Retrieval via Multi-Modal Entailment}

\author{
    Xu Yan\textsuperscript{\rm 1}
    Chunhui Ai\textsuperscript{\rm 1}
    Ziqiang Cao\textsuperscript{\rm 1}
    Min Cao\textsuperscript{\rm 1}\\
    Sujian Li\textsuperscript{\rm 2}
    Wenjie Li\textsuperscript{\rm 3}
    Guohong Fu\textsuperscript{\rm 1}
}
\affiliations{
    \textsuperscript{\rm 1}Soochow University,
    \textsuperscript{\rm 2}Peking University,
    \textsuperscript{\rm 3}Hong Kong Polytechnic University\\

    \{20215227006, 20215227120\}@stu.suda.edu.cn, 
    \{zqcao, mcao, ghfu\}@suda.edu.cn,\\
    lisujian@pku.edu.cn,
    cswjli@comp.polyu.edu.hk
}

\usepackage{bibentry}

\begin{document}

\maketitle

\begin{abstract}

An outstanding image-text retrieval model depends on high-quality labeled data. 
While the builders of existing image-text retrieval datasets strive to ensure that the caption matches the linked image, they cannot prevent a caption from fitting other images. 
We observe that such a many-to-many matching phenomenon is quite common in the widely-used retrieval datasets, where one caption can describe up to 178 images. 
These large matching-lost data not only confuse the model in training but also weaken the evaluation accuracy. 
Inspired by visual and textual entailment tasks, we propose a multi-modal entailment classifier to determine whether a sentence is entailed by an image plus its linked captions. 
Subsequently, we revise the image-text retrieval datasets by adding these entailed captions as additional weak labels of an image and develop a universal variable learning rate strategy to teach a retrieval model to distinguish the entailed captions from other negative samples. 
In experiments, we manually annotate an entailment-corrected image-text retrieval dataset for evaluation. 
The results demonstrate that the proposed entailment classifier achieves about 78\% accuracy and consistently improves the performance of image-text retrieval baselines.

\end{abstract}

\section{Introduction}


Image-text retrieval aims to retrieve items through visual or semantic information.
It contains two sub-tasks: image retrieval and text retrieval, depending on which modality is used as the retrieved target. 
Image-text retrieval has been widely adopted in various applications, such as the retrieval of commodity pictures given textual descriptions.
Most image-text retrieval approaches \citep{li_visualbert_2019,li_unicoder-vl_2019,tan_lxmert_2019,li_visual_2019,su_vl-bert_2020} focus on mapping features of image and text modalities into a common semantic space.
Notably, recent studies \citep{li_oscar_2020,chen_uniter_2020,jia_scaling_2021,radford_learning_2021,li_align_2021} have shown that Vision-and-Language Pre-training  (VLP)  can effectively learn general representations and achieves high performance on this task.

\begin{figure}
	\small
    \centering
	\includegraphics[width=0.9\linewidth]{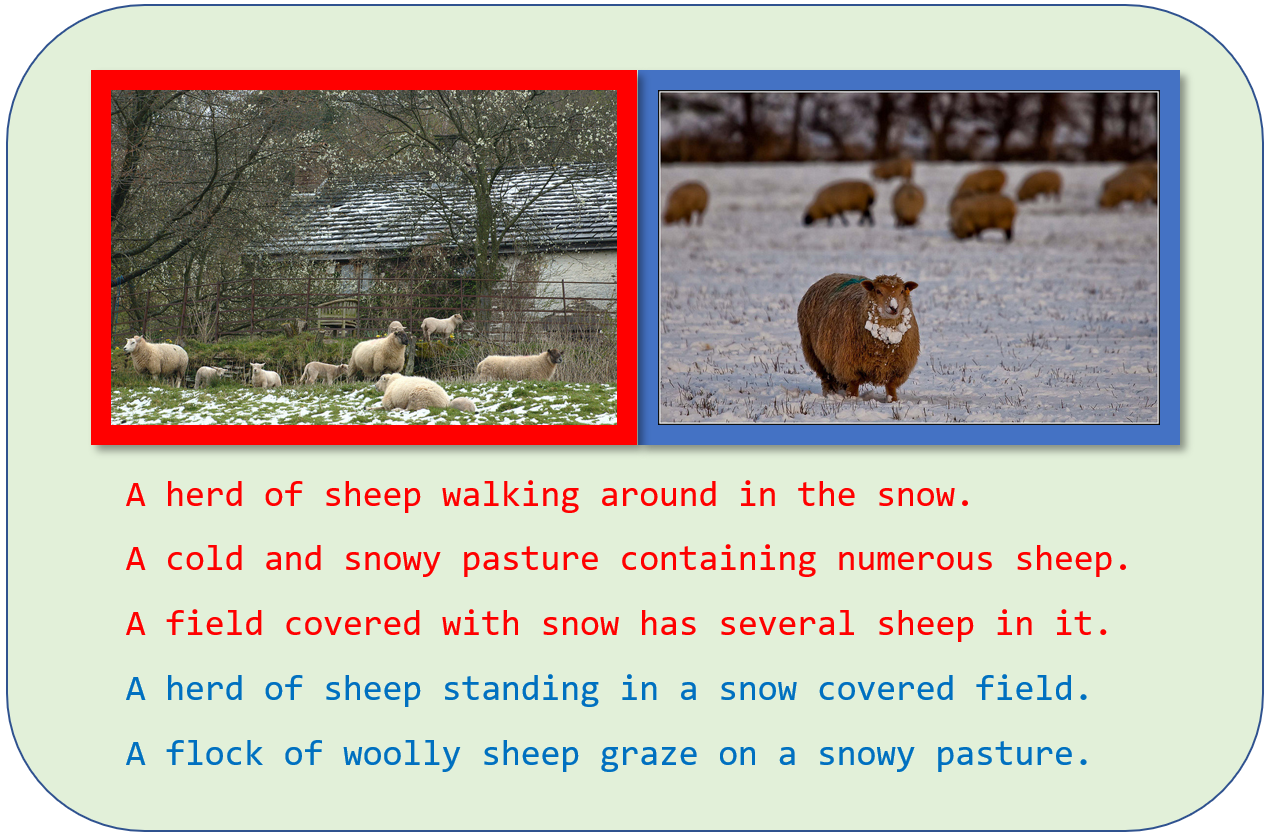} 
	\caption{
	Examples of images and texts from MSCOCO dataset.
    While all of captions can describe the two images,
    only image-text pairs with the same color are marked as positive pairs.
	}
	\label{Fig.main2} 
\end{figure}

Image-text retrieval relies on curated training datasets that are usually expensive and sometimes even require expert knowledge to acquire.
Common image-text retrieval datasets, including Flickr8K~\citep{rashtchian_collecting_nodate}, Flickr30K~\citep{young_image_2014}, Multi30k~\citep{elliott_multi30k_2016} and MSCOCO~\citep{lin_microsoft_2015}, are constructed through manually writing a few descriptive captions for each image using crowd-sourcing.
Therefore, it is only ensured that the image and its descriptive captions are matched when annotated.
However, the possible associations between an image and other captions in the dataset are not fully considered.
Taking Figure~\ref{Fig.main2} as an example, two images depicting the same scene have their different text descriptions, which can also be used to describe each other.
Such a many-to-many matching phenomenon is quite common in retrieval datasets.
For example, in MSCOCO, we find that 89 captions can describe one image while this number amazingly reaches 178 on the text side (refer to Section~\ref{section:annotation} for more details).
Unfortunately, the cross-matched image-text pairs with similar semantics are typically regarded as negative examples.
As we know, treating semantically matched image-text pairs as negative in training will increase their distance in vector space and thus reduce the quality of representation learning.
Meanwhile, marking them as errors in evaluation leads to a significant false negative rate.

This paper proposes an automatic solution to handle the many-to-many matching problem in the retrieval datasets.
Our solution recognizes this kind of relationship and utilizes the relationship in training.
We argue that if an image and its descriptive captions entail the meaning of a sentence, this sentence should be able to describe the image.
Inspired by the tasks of visual entailment \citep{xie_visual_2019} and textual entailment \citep{snli}, we propose a multi-modal entailment classifier to recognize the entailment relationship between a caption and an image combined with its descriptive captions.
To fully utilize the external textual and visual entailment data, our entailment model supports various forms of input, including text-text, image-text, and image\&text-text.
We modify existing models~\citep{li_align_2021,devlin_bert_2019} to conduct textual entailment and visual entailment, and combine the hidden states of textual/visual modules to produce the final multi-modal entailment result.
Next, we use this entailment model to find the entailed image-text pairs in the retrieval datasets.
During training, we treat these entailed pairs as additional weak positive samples and set a small learning rate for them.
This learning strategy can be used for any retrieval model without changing its internal structure.

In order to verify the proposed entailment model, we manually annotated an entailment-corrected dataset containing 2k image-text pair samples from MSCOCO and Flickr30K.
Results show that our entailment classifier achieves about 78\% accuracy.
Moreover, trained on image-text pairs revised by our entailment classifier, the retrieval models uniformly achieve a performance improvement in both retrieval and entailment evaluations.

The contributions of this paper can be summarized as follows:
\begin{itemize}
    \item We utilize multi-modal entailment to handle the many-to-many matching problem in image-text retrieval datasets and annotate an entailment-corrected dataset for evaluation\footnote{Code and the dataset will be released in the final version.}.
    \item We propose a strong multi-modal entailment classifier to determine the entailed image-text pairs in the retrieval datasets automatically.
    \item We develop a universal entailment-enhanced learning strategy to consistently to improve retrieval models' matching performance consistently.
\end{itemize}

\section{Related Work}

\subsection{Image-Text Retrieval Datasets}

Early image-text datasets include Flickr8K \citep{rashtchian_collecting_nodate} and Flickr30K \citep{young_image_2014}. 
Inspired by them, \citet{lin_microsoft_2015} builds a larger Microsoft Common Objects in COntext (MSCOCO) Caption dataset. 
A number of datasets subsequently emerge such as Multi30k \citep{elliott_multi30k_2016}, Conceptual Captions \citep{sharma_conceptual_2018} and RedCaps \citep{RedCaps}.
Notably, Conceptual Captions and RedCaps are built through web crawling, while others are constructed by manually writing a few descriptive captions for each image using crowd-sourcing.
All these datasets only ensure relationships between images and texts created for them and ignore possible associations of external image-text pairs.

Some recent works have been aware of this problem and attempted to introduce many-to-many correspondences for image-text datasets.
CrissCrossed Caption (CxC) \citep{parekh_crisscrossed_2021} and Extended COCO Validation (ECCV) \citep{chun_eccv_2022} datasets are built through  manually annotating sampled MSCOCO image-text pairs with similarity scores or categories.
However, due to expensive labor costs and unscalable annotations, it is challenging to construct a large-scale dataset for training. 
Moreover, the human similarity score does not entirely fit the retrieval task, and even image-text pairs with high scores cannot always be taken as positive samples.
For example, in the CxC dataset, the caption ``A couple of birds that are walking on some sand.'' matches the image with a single seagull.

\subsection{Textual Entailment and Visual Entailment}
Textual entailment \citep{Dagan2005ThePR}, often used as a benchmark to measure the ability of language understanding \citep{Dagan2005ThePR,bowman-etal-2015-large}, has been a hot research topic in the NLP area.
In the last few years, with the advancement of deep learning, the study of textual entailment is gradually being carried out on some large-scale data such as SNLI \citep{DBLP:journals/corr/BowmanAPM15}, SciTaiL \citep{Khot2018SciTaiLAT}, MNLI \citep{DBLP:journals/corr/WilliamsNB17}, and XNLI \citep{conneau2018xnli}.
In addition, textual entailment in the context of the few-shot scenario has also been much studied, like UFO-ENTAIL \citep{DBLP:journals/corr/abs-2010-02584}. 

Inspired by textual entailment, \citet{DBLP:journals/corr/abs-1901-06706} proposes visual entailment task to determine the entailment between a given image and text pair.
They annotate a dataset SNLI-VE by linking SNLI to Flickr30K.
In recent studies, it has often been treated as a downstream task of Vision-and-Language Pre-training(VLP) model \citep{huang2021seeing,albef,DBLP:journals/corr/abs-2108-10904,DBLP:journals/corr/abs-2202-03052}.
In addition, \citet{ilharco-etal-2021-recognizing} proposes a multi-modal entailment dataset, but the dataset is not well adapted to our multi-modal entailment model.


\section{Multi-Modal Entailment Classifier}

\begin{figure*}[htb]
    \centering
    \includegraphics[width=0.8\linewidth]{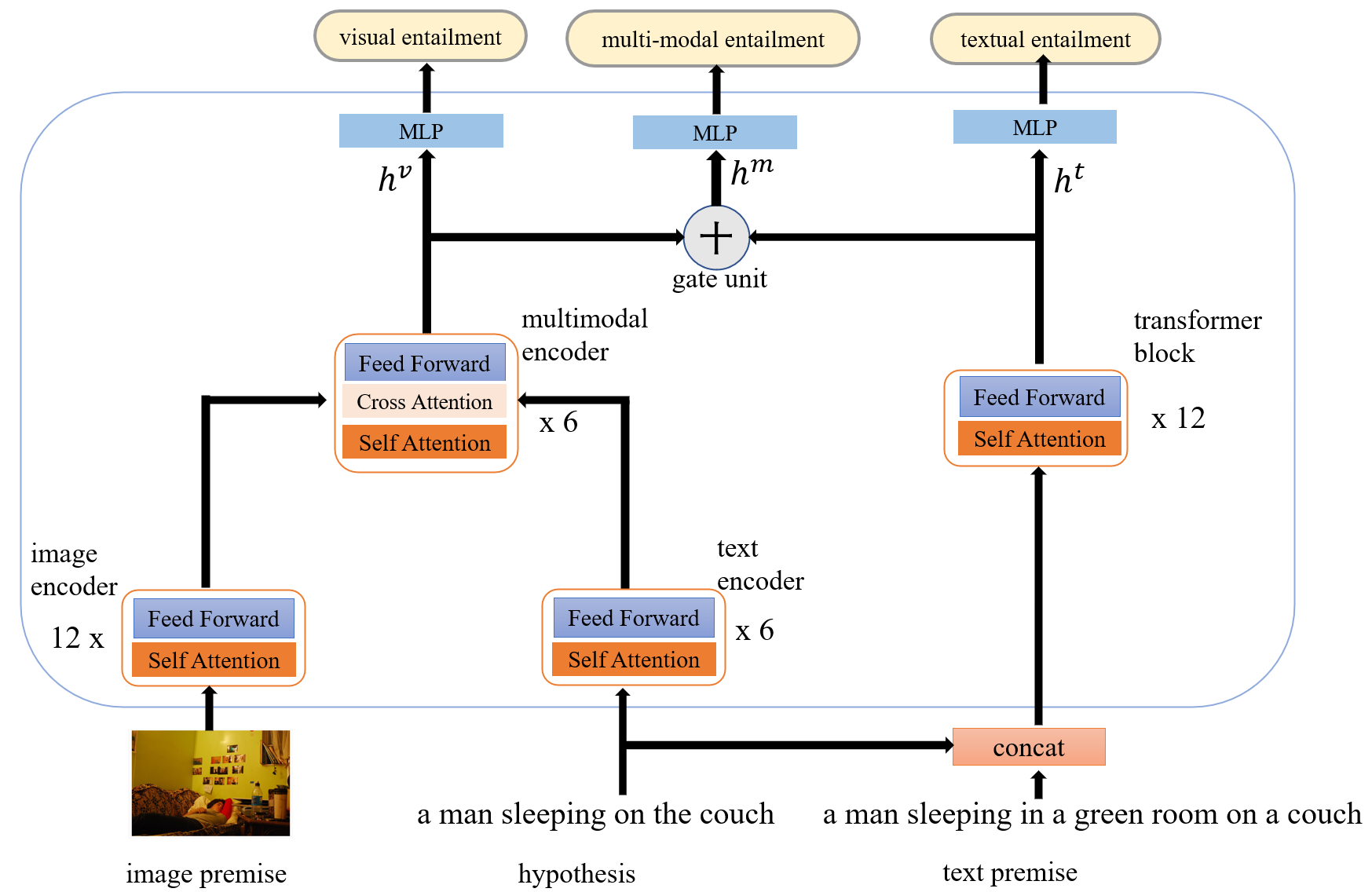}
    \caption{Illustration of our multi-modal entailment classifier.
    It consists of a visual entailment module and a textual entailment module.
    The result of multi-modal entailment is obtained by combining the hidden states of visual and textual entailment through a gate unit.}
    \label{mme}
\end{figure*}

The proposed multi-modal entailment classifier is used to recognize whether a sentence is entailed by an image plus its captions.
We utilize the classifier to construct the entailment-revised retrieval dataset for training automatically.
Figure~\ref{mme} shows the model structure.
It contains a visual entailment module and a textual entailment module and combines the hidden states of the two modules to predict the final multi-modal entailment category. 
Our model supports three types of input premises: an image, text, and a combination of image and text.
Note that to be adaptable to downstream image-text retrieval tasks, we only classify the relationship into entailment or non-entailment, rather than the traditional entailment task with three categories: entailment, neutral, and contradiction.
In the following description we use $x^{p_v}$ and $x^{p_t}$ for the image and text in  premise, $x^h$ for the text hypothesis and $y\in\{0,1\}$ for the target where 1 means entailment and 0 means non-entailment.
This section will illustrate how our model conducts the three types of entailment data.

\subsection{Textual Entailment}\label{te}

In textual entailment, both the premise and hypothesis are textual sentences, namely the $\text{input}=(x^{p_t},x^h)$. We define this form of the task as text-text and adopt BERT \citep{devlin_bert_2019} as our backbone model.

Following the common practice, we pack two sentences $x^{p_t}$ and $x^h$ together as $([cls],x^{p_t},[sep],x^h)$, where $[cls]$ and $[sep]$ are two special tags.
Next, the packed texts are fed into the BERT model to get the entire representation:
\begin{equation}
    h^t = BERT(x^{p_t},x^h).
\end{equation}
Like \citet{DBLP:journals/corr/abs-2101-10642}, we just use the hidden state at the sentence tag ($[cls]$) to represent the entire input.
On top of $h_t$, we add a simple multi-layer perceptron (MLP) classifier with two hidden layers to predict the final label:
\begin{equation}
    p(\hat{y}|x^{p_t},x^h) = softmax(MLP(h^t)).
\end{equation}
where we adopt ReLU \citep{pmlr-v15-glorot11a} as the activation function for MLP.
Notably, we use softmax rather than sigmoid for this binary classification task as we compare the two methods, and the results show that softmax is $1.8\%$ higher than sigmoid.

\subsection{Visual Entailment}\label{ve}
In visual entailment, the premise is an image $x_{p_v}$, and the task form is defined as image-text.
We adopt the structure of the state-of-art image-text retrieval model ALBEF \citep{albef} to encode $x_{p_v}$ and $x_h$, namely:
\begin{equation}
    h^v = ALBEF(x^{p_v},x^h).
\end{equation}
ALBEF consists of a 12-layer visual transformer (ViT) ~\citep{vit} as the image encoder and a 6-layer transformer for both text encoder and multi-modal encoder.
The cross-attention mechanism in a multi-modal encoder achieves an alignment between visual and textual modals.
Similar to textual entailment, after a simple multi-layer perceptron with two hidden layers, we can get a distribution of prediction $\hat{y}$.
\begin{equation}
    p(\hat{y}|x^{p_v},x^h) = softmax(MLP(h^v)).
\end{equation}

Referring to the practice of \citet{liang_not_all_patch_2022} in ViT, we develop an image augment method to increment negative samples.
Concretely, ViT will split an image into patches and encode them by self-attention mechanism \citep{vaswani_attention_2017}.
Intuitively, patches with higher attention scores should represent more significant regions and play a critical role in recognizing entailment relationships.
For images of positive samples, we mask their partial patches with the highest score according to the attention matrix in ViT.
Through this augment, original image-text pairs will become non-entailment and supply negative samples.
In the experiments, the masking ratio is a hyper-parameter we set as 0.4, and in each batch, we select up to 4 images for mask augment.

\subsection{Multi-Modal Entailment} \label{me}
In textual entailment and visual entailment, the premise is just uni-modal.
However, we actually need to check whether a sentence is entailed by an image plus its captions, and we define the form of the task when the premise input of our task is multi-modal as image\&text-text.
In this section, we want to combine textual and visual entailment for multi-modal entailment.
The data pairs are defined as $(x^{p_v}+x^{p_t},x^h)$.
Briefly, we merge the captions of the same image to form $x^{p_t}$.
Inspired by \citet{xu-etal-2021-read}, we want to build a gate unit to combine visual entailment and textual entailment to make a comprehensive judgment.
Given the hidden states $h^t$ and $h^v$ computed in the above textual entailment and visual entailment modules, we propose a gate unit to merge them into multi-modal hidden states:
\begin{align}
    g^t &= \sigma (W^th^t+b^t), \\
    g^v &= \sigma (W^vh^v+b^v), \\
    h^m &= g^t \cdot h^t + g^v \cdot h^v.
\end{align}
where $W^t$, $b^t$, $W^v$, $b^v$ are learnable parameters and $\sigma$ is sigmoid function.
Finally, the classification is done by a multi-layer perceptron classifier with two hidden layers:
\begin{equation}
    p(\hat{y}|x^{p_v},x^{p_t},x^h) = softmax(MLP(h^m)).
\end{equation}

We have tried to merge $x^{p_v}$, $x^{p_t}$ and $x^h$ directly using a multi-modal encoder instead of a gate unit, but this can easily cause memory overflow and make it impossible to separate visual and textual entailment.

\subsection{Joint Learning}

The learning process is driven by optimizing three objectives, corresponding to visual entailment $\mathcal{L}_{v}$, textual entailment $\mathcal{L}_{t}$ and multi-modal entailment $\mathcal{L}_{m}$ respectively.

\begin{align}
    \mathcal{L}_{t} &= - \sum_{i} \log p(\hat{y_i} = y_i | x_i^{p_t}, x_i^h), \\
    \mathcal{L}_{v} &= - \sum_{i} \log p(\hat{y_i} = y_i |x_i^{p_v}, x_i^h), \\
    \mathcal{L}_{m} &=  - \sum_{i} \log p(\hat{y_i} = y_i |x_i^{p_v}+x_i^{p_t}, x_i^h).
\end{align}

To facilitate training, we unify the input form of the model as the multi-modal task.
To achieve this goal, we fill plain black images for textual entailment and empty premise strings for visual entailment.
Meanwhile, we introduce three binary indicators $\theta _v,\theta _t,\theta _m$ to accumulate the related losses for backpropagation:
\begin{equation}
    \mathcal{L_{\text{all}}} = \theta _t \mathcal{L}_{t} + \theta _v \mathcal{L}_{v} + \theta _m \mathcal{L}_{m}.
\end{equation}
For textual entailment, only $\theta _t=1$ and for visual entailment, only $\theta _v=1$, while all the losses are used in multi-modal entailment. 

\begin{figure}[htb] 
    \centering 
    \includegraphics[width=0.48\textwidth]{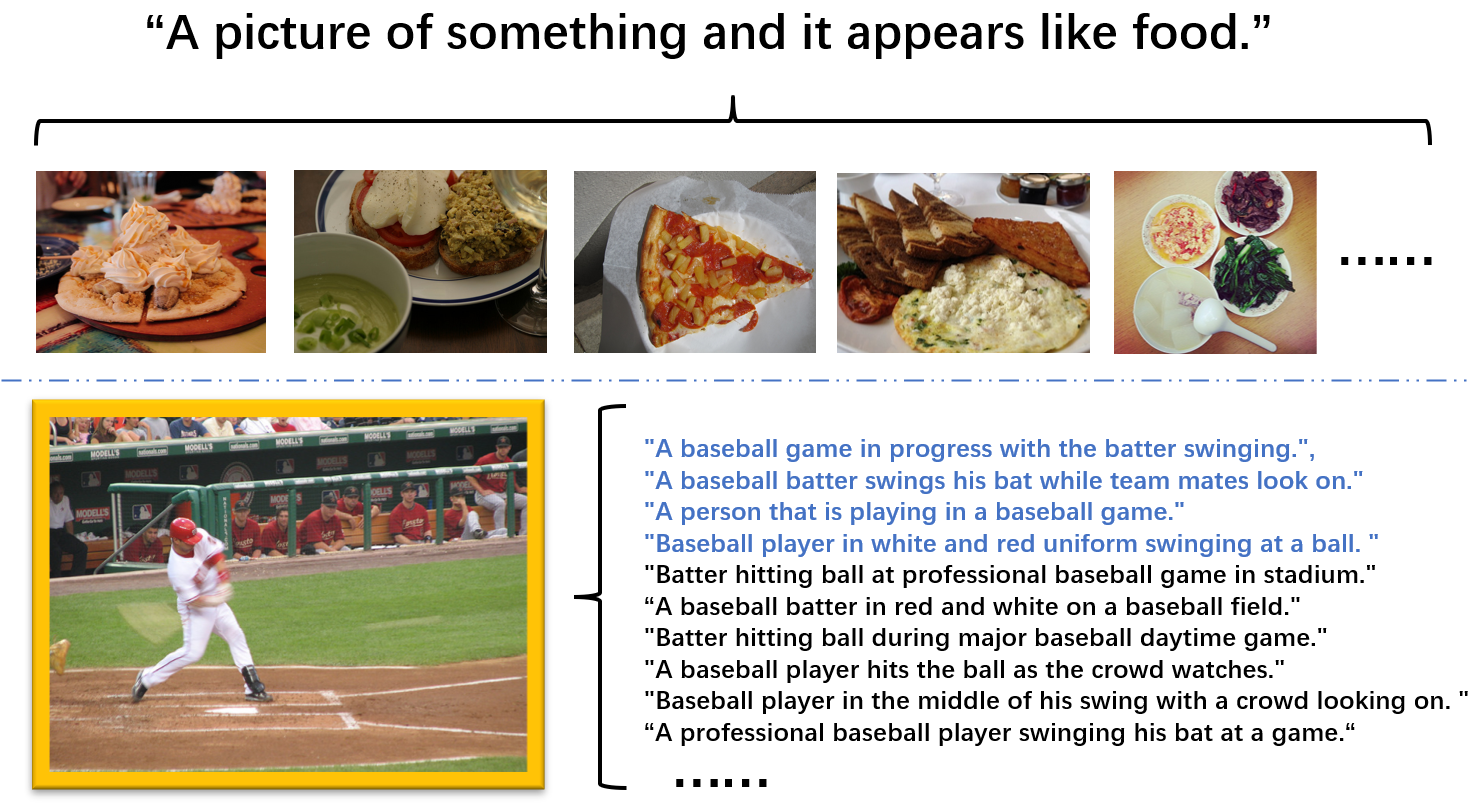} 
    \caption{Typical examples about how many items that one image or caption can match.
    Blue: original positives.
    } 
    \label{Fig.maxmatch} 
\end{figure}

\section{Entailment-Enhanced Training for Retrieval Models}

We automatically detect the entailed image-text pairs in image-text retrieval datasets with the proposed multi-modal entailment classifier.
Subsequently, we use entailed pairs in the following two aspects.
On the one hand, current image-text retrieval models usually adopt negative sampling \citep{li_align_2021,radford_learning_2021,chen_uniter_2020} to enforce dissimilar representations between non-golden image-text pairs.
In the training process, we optimize the negative sampling method by preventing captions from being selected as negative samples for entailed images.
On the other hand, we regard these extra entailed image-text pairs as weak positives and propose a universal variable learning rate strategy to handle them.
Expressly, assume that the learning rate of the golden positive examples during training is $\lambda$.
Then we apply a smaller learning rate $\lambda^{'}$ to weak positives, where $\lambda^{'} = \alpha\lambda$ and $\alpha \in (0,1)$ is a hyper-parameter.

In subsequent experiments, we empirically set $\alpha$ to $0.3$.
Considering the learning rate cannot be distinguished within the same batch, we assemble weak positives into an additional batch immediately after each regular batch.
We preferentially select weak positives according to images in the regular batch.

These two methods above allow semantically related images and texts to be close to each other without introducing too much noise in training.
While optional methods include contrastive learning \citep{Noise-contrastive} and applying different weights on training loss for weak positives, they need to modify models specifically.
They are not as universal as our strategy.
Our experiments show that our methods can effectively enhance the entailment degree of the retrieval models while keeping or improving the retrieval performance.

\section{Entailment-Corrected Dataset Annotation} \label{section:annotation}

\begin{table}[tp]
  \centering
  \small
    \begin{tabular}{c|c|c}
    \toprule
          & Flickr30K & MSCOCO \\
    \midrule
    Total pairs & 1000 & 1000 \\
    Entailment & 699 & 307 \\
    \bottomrule
    \end{tabular}%
  \caption{Statistics of the entailment-corrected dataset.}
  \label{tab:testset}%
\end{table}%

We manually annotate an entailment-corrected dataset to evaluate the effects of our multi-modal entailment model.
We select images and texts from the MSCOCO and Flikr30K test datasets to improve their diversity.

Since most of the image-text pairs in retrieval datasets are semantically irrelevant and have no entailment relationship, we use a fine-tuned retrieval model ALBEF to get the top-30 text retrieval results as annotation candidates.
After sampling images in the candidates, we randomly select one text for every image.
In this way, the assembled image-text pairs usually hold high semantic association.
We also add a small part of random image-text pairs to ensure the diversity of our dataset.

Seven graduate students are arranged for annotation.
They must make an inference for the hypothesis sentence according to the given premise.
To better use multi-modal information for entailment relationship classification, every premise in our dataset includes both image and its linked ground truth captions.
More details of our dataset are shown in Appendix A.
A hypothesis sentence can be regarded as entailment with its premise only if it meets the following two points:
\textbf{(1)} This hypothesis sentence must clearly describe the content of the image premise without ambiguity.
\textbf{(2)} This hypothesis sentence can be inferred from premise texts and cannot be contradictory to them all.
All pairs not meeting the above conditions are regarded as negative examples.
Testing on 30 identical samples, the Kappa score~\citep{Fleiss_kappa_2015} of annotators reaches about $0.8$, indicating high consistency.
Finally, we get 1k labeled image-text pairs for Flickr30K and 1k for MSCOCO.
Statistics about our dataset are shown in Table~\ref{tab:testset}.

In addition, we use the same method to annotate some typical examples in the original MSCOCO testset.
As shown in Figure~\ref{Fig.maxmatch}, we found that one plain caption ``A picture of something and it appears like food'' can match accord with up to 178 images with food, and the image with a person who is playing a baseball game can be depicted by according up to 89 captions.
These huge numbers demonstrate the universality of the many-to-many matching phenomenon.
We also find contradictions even in the original golden image-text pairs. 
For example, different annotators describe a child in the same picture as a boy and a girl.

\section{Experiment}

In this section, we present experimental results for our multi-modal entailment classifier and the proposed entailment-enhanced training for various retrieval models.

\subsection{Datasets}

\paragraph{Multi-Modal Entailment}

\begin{table}[tb]
\centering
\small
\resizebox{0.45\textwidth}{!}{
\begin{tabular}{c|c|l|l}
\cline{1-4}
 & \multicolumn{1}{l|}{Task }                    & Dataset                  & Count  \\ \cline{1-4}
\multirow{12}{*}{Train} & \multirow{6}{*}{TE}       & XNLI \citep{conneau2018xnli}               & 400.2k  \\ \cline{3-4}
                        &                                           & MRPC \citep{dolan-brockett-2005-automatically}               & 5.8k       \\ \cline{3-4}
                        &                                           & RTE \citep{bentivogli_fifth_nodate}                & 2.7k       \\ \cline{3-4}
                        &                                           & STS-B  \citep{cer-etal-2017-semeval}            & 7.2k       \\ \cline{3-4}
                        &                                           & QQP  \citep{Chen2017QuoraQP}              & 404.2k       \\ \cline{3-4}
                        &                                           & TS \citep{kauchak-2013-improving}                & 167.6k       \\ \cline{2-4}
                        &   \multirow{2}{*}{VE}    & SNLI-VE \citep{DBLP:journals/corr/abs-1901-06706}           & 529.5k       \\ \cline{3-4}
                        &                                           & Image Masking                & 132.3k       \\ \cline{2-4}
                        & \multirow{4}{*}{MME} & SNLI-VE           & 529.5k       \\ \cline{3-4}
                        &                                           & CXC \citep{parekh_crisscrossed_2021}               & 39.5k       \\ \cline{3-4}
                        &                                           & ECCV  \citep{chun_eccv_2022}             & 26.4k       \\ \cline{3-4}
                        &                                           & Image Masking              & 148.8k       \\ \cline{1-4}
Dev                     & \multicolumn{1}{l|}{}                     & SNLI-VE     & 17.8k       \\ \cline{1-4}
Test &
\multirow{1}{*}{}    &                    Annotated Dataset & 2k       \\ \cline{1-4}
\end{tabular}}
\caption{Statistics of datasets used in the multi-modal entailment task. TE, VE, and MME denote textual entailment, visual entailment, and multi-modal entailment, respectively.}
\label{mmed}
\end{table}

\begin{table}[tp]
\centering
\small
\resizebox{0.48\textwidth}{!}{
\begin{tabular}{l|cccc}
\hline
\textbf{Model}   & accuracy   & precision   & recall   & $f_{0.5}$
\cr
\hline
Only TE & 71.1 & 65.0 & 90.9 & 68.9                        
\cr \hline 
Only VE  & 72.3 & 66.9 & 87.5 & 70.2                       
\cr \hline 
OFA                     & 73.3 & 67.4 & 89.6 & 70.9                        
\cr \hline 
Ours                    & 78.1 & 80.2 & 74.3 & \textbf{78.9}        
\cr 
w/o Image Masking  & 78.4 & 77.7 & 79.4 & 78.0  
\cr
w/o VE Data  & 66.4 & 62.5 & 81.9 & 65.6  
\cr 
w/o TE Data & 77.7 & 74.2 & 84.6 & 76.1
\cr
w/o BERT & 76.5 & 72.4 & 85.1 & 74.6
\cr \hline  
\end{tabular}}
\caption{Performance (\%) of different entailment models tested on our annotated dataset.
w/o BERT means using a text encoder from ALBEF in the textual entailment.}
\label{results_mme}
\end{table}

The datasets we used for textual entailment, visual entailment, multi-modal entailment are listed in Table~\ref{mmed}.
More details of these datasets are described in Appendix B.
For visual entailment, we perform image data augment by masking critical patchs of images, as described in Section~\ref{ve}.
In addition, we try to use golden captions in our datasets as data augmentation.
Specifically, we randomly select four captions from the corresponding five captions of each image as textual premises and the rest as hypotheses to construct implicit positive samples.
However, experimentally we find that this data augmentation method reduces the model's generalization ability.

\paragraph{Image-Text Retrieval}

We consider two widely-used datasets for image-text retrieval tasks: MSCOCO and Flickr30K. 
Specifically, we adopt both datasets' widely used Karpathy split \citep{Karpathy_2015}.
The MSCOCO contains 113/5k/5k for train/validation/test, and
the Flickr30K contains 29k/1k/1k images for train/validation/test. 
We present experimental results on MSCOCO 5K and Flickr 1K testsets.

\subsection{Baseline Models}
\begin{table*}[htp]
  \centering
    \small
    \begin{tabular}{l|cccccc}
    \toprule
    \multirow{2}[6]{*}{\textbf{Method}} & \multicolumn{6}{c}{Flickr30K / MSCOCO}\\
    \cmidrule{2-7}  & TR@1.  & TR@5.  & TR@10. & IR@1.  & IR@5.  & IR@10. \\
    \midrule
    
    ALBEF   & 95.2 / 77.4 & 98.9 / 93.9 & 100.0 / 97.1  & 85.3 / 61.2 & 97.3 / 84.6 & 98.7 / 91.0  \\

    ALBEF\(^{\#}\)  & +0.1 / +0.2 & +0.6 / +0.2 & -0.2 / +0.2  & +0.2 / -0.3 & +0.1 / -0.1 & 0.0 / -0.1 \\

    \midrule
    
    CLIP   & 89.2 / 64.5 & 97.4 / 85.9 & 99.4 / 92.2  & 74.4 / 47.4 & 93.5 / 74.4 & 96.7 / 83.4  \\

    CLIP\(^{\#}\)  & \textbf{+1.6} / \textbf{+2.0} & \textbf{+1.4} / \textbf{+1.1} & +0.5 / +0.5  & \textbf{+3.1} / \textbf{+1.5} & \textbf{+2.1} / \textbf{+1.4} & +0.9 / \textbf{+1.0}  \\
    
    \midrule
    
    UNITER   & 84.2 / 64.7 & 97.1 / 88.2 & 98.7 / 93.5  & 70.8 / 49.1 & 91.7 / 77.4 & 95.5 / 86.0  \\

    UNITER\(^{\#}\) & \textbf{-1.0} / +0.4 & +0.1 / \textbf{+1.3} & +0.1 / 0.0  & +0.4 / \textbf{+1.3} & +0.6 / +0.1 & +0.7 / +0.9  \\
    \bottomrule
    \end{tabular}

  \caption{Performance (\%) of different image-text retrieval models finetuned on Flickr30K and MSCOCO.
  The scores before and after the symbol "/" represent the evaluation results on original Flickr30K and MSCOCO testsets, respectively.
  "\(^{\#}\)" denotes the model is trained with our entailment-enhanced strategy.
  The changes $\ge 1.0$ are shown in \textbf{bold}.
  }
  \label{tab:F30kandCOCO}
\end{table*}

\paragraph{Multi-Modal Entailment}

We adopt BERT \citep{devlin_bert_2019} and ALBEF \citep{li_align_2021} as the backbone structure of textual entailment and visual entailment.
Therefore we test the performance using each module.
In addition, we introduce OFA \citep{DBLP:journals/corr/abs-2202-03052}, a state-of-the-art visual entailment classifier, as a comparison baseline.

\paragraph{Image-Text Retrieval}

We compare our variable learning rate strategy with some competitive image-text retrieval models, including ALBEF, CLIP \citep{radford_learning_2021} and UNITER \citep{chen_uniter_2020}.
More details of these baseline models are described in Appendix C.

\subsection{Evaluation Metrics}

\paragraph{Multi-Modal Entailment}

The accuracy, precision, and recall of our annotated dataset are reported as the evaluation metrics, which are commonly used in the entailment task.
Particularly, following the \citet{zhang_overview_2018},
We put more weight on precision and apply $F_{0.5}$ as our final evaluation metric.

\paragraph{Image-Text Retrieval}

As the common practice \citep{Karpathy_2015}, we report the \textbf{Recall@K} (R@K) as evaluating metrics, which measures the fraction of times a correct item was found among the top K results.
For text-retrieval (TR) and image-retrieval (IR), we report TR@1/5/10 and IR@1/5/10, respectively.

To quantitatively measure the relevance between retrieved texts and the query images, we propose a novel metric called \textbf{Entail@K} (E@k).
E@K measures the averaged entailment ratio in the top-k retrieved items:
\begin{equation}
    Entail@K = \frac{1}{K}\sum_{i=1}^{K}e_{i}(x),
\end{equation}
where the binary indicator $e_{i}(x)$ equals $1$ If and only if the $i$-th retrieved text is ground truth or has an entailment relationship with the query image $x$.
Higher E@k values mean that the retrieved texts have a stronger descriptive and semantic association with the query images.

For the image-text pairs included in our entailment-corrected dataset, the relationship can be obtained directly.
For the rest pairs, we use two ways to get their entailment labels.
On the one hand, we sample some images and manually annotate the entailment relationship of their retrieval results with the same rules as Section~\ref{section:annotation}.
On the other hand, we use our trained multi-modal entailment model to infer the relationship between image $x$ and $i$-th text.
The manual method is more accurate but requires too much cost, while the automatic way can quickly evaluate all the datasets.

In subsequent experiments, we randomly selected 50 common images with their retrieved top-10 texts from the text-retrieval results on both testset of Flickr30K and MSCOCO for manual annotation.
We denote these manual entailment results with \textbf{E@M}.

\subsection{Implementation Details}

We mix the textual, visual, and multi-modal entailment data and train them together indiscriminately for our multi-modal entailment model.
We found that this mixing strategy is much better than training separately.
We trained the multi-modal entailment model with five epochs on 8 Amax-5000 GPUs with a batch size of 96.
We use the AdamW \citep{loshchilov_adamw_2019} optimizer with a weight decay of 0.02 and initial learning rate 2e-5.

For image-text retrieval, due to models' scales, we set different batch sizes and initial learning rates for different models (i.e., 96/2e-5 for ALBEF, 1536/1e-5 for CLIP, 96/5e-5 for UNITER).
We use the AdamW optimizer with a weight decay of 0.02.

\begin{table}[tp]
  \centering
  \small
    \begin{tabular}{l|c|c|c}
    \toprule
    Method          & E@10 & E@30  & E@M \\
    \midrule
    ALBEF           & 63.9 & 44.1  & 76.7 \\
    ALBEF\(^{\#}\)  & \textbf{66.0} & \textbf{46.0}  & \textbf{78.0} \\
    \midrule
    CLIP            & 58.2 & 41.5  & 67.4\\
    CLIP\(^{\#}\)   & \textbf{60.9} & \textbf{43.2}  & \textbf{75.5} \\
    \midrule
    UNITER          & 44.9 & 27.2  & 73.1\\
    UNITER\(^{\#}\) & \textbf{46.9} & \textbf{28.5}  & \textbf{76.4} \\
    \bottomrule
    \end{tabular}%
  \caption{Performance of E@k on different retrieval models.
 E@M  stands for evaluation by manually annotated 50 common samples.
 E@10/30 are averaged scores over Flickr30K and MSCOCO testsets.
 }
  \label{tab:manual entailment}%
\end{table}%

\subsection{Main Results}

\begin{figure*}[htb] 
    \centering 
    \includegraphics[width=1\textwidth]{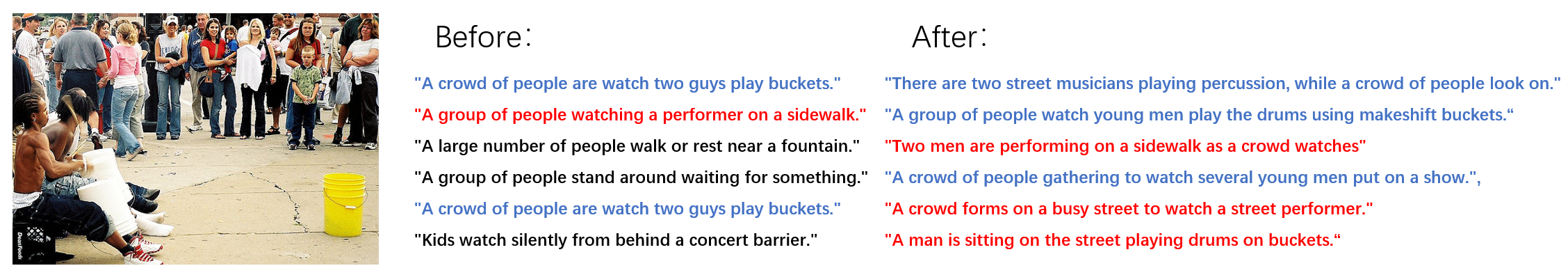} 
    \caption{Comparison of examples of retrieval results before and after applying our entailment-enhanced learning strategy.
    Blue: original positives.
    Red: manually annotated entailment samples.
    Black: irrelevant samples.
    } 
    \label{Fig.case4} 
\end{figure*}

\subsubsection{Multi-Modal Entailment}

The results of the entailment experiments are shown in Table~\ref{results_mme}.
As can be seen, our multi-modal entailment model all the other baselines to a large extent.
For instance, the $f_{0.5}$ is more than 8\% larger than the state-of-the-art visual entailment model OFA.
The results demonstrate our proposed multi-modal entailment model is more competitive than the traditional textual and visual entailment models.
Meanwhile, the precision of annotated dataset has improved dramatically, which guarantees the possibility that the model will be used for automatic detection.
In addition, we conduct a series of ablation experiments for training data.
As can be seen, removing any training data will degrade the f-score, while the labeled visual entailment data seem more critical.
A possible reason is that the visual entailment datasets fit the multi-modal entailment task well.
We use the text encoder from ALBEF as a comparison, and the results show that the $f_{0.5}$ was about 4.3\% higher using BERT.
Overall, both the textual and visual entailment modules are helpful, making an essential contribution to our model in learning more about multi-modal interactions.

\subsubsection{Entailment-Enhanced Training Strategy}

Table~\ref{tab:F30kandCOCO} shows the results of different retrieval methods with or without applying our variable learning rate strategy on two benchmarks, Flickr30K and MSCOCO, respectively.
Although we focus on the improvement of many-to-many matching recognition, we find that our entailment-enhanced training could also often improve the retrieval performance.
Especially for CLIP's IR@1 score on Flickr30K raises more than 3\% with our learning strategy. 
Therefore, we believe our entailment-enhanced training indeed helps the retrieval models find appropriate positive and negative image-text pairs.

In addition, we demonstrate the entailment performance of different retrieval models in  Table~\ref{tab:manual entailment}.
As can be seen, after applying our entailment-enhanced training strategy, all models' entailment performance obviously improves on both automatic and manual evaluations.
Notably, CLIP$^\#$ significantly exceeds CLIP by more than 8\% in terms of E@M.
The results reveal the effectiveness of our strategy in refining the entailment degree for retrieval models universally.

\begin{figure}[htb] 
    \centering 
    \includegraphics[width=0.48\textwidth]{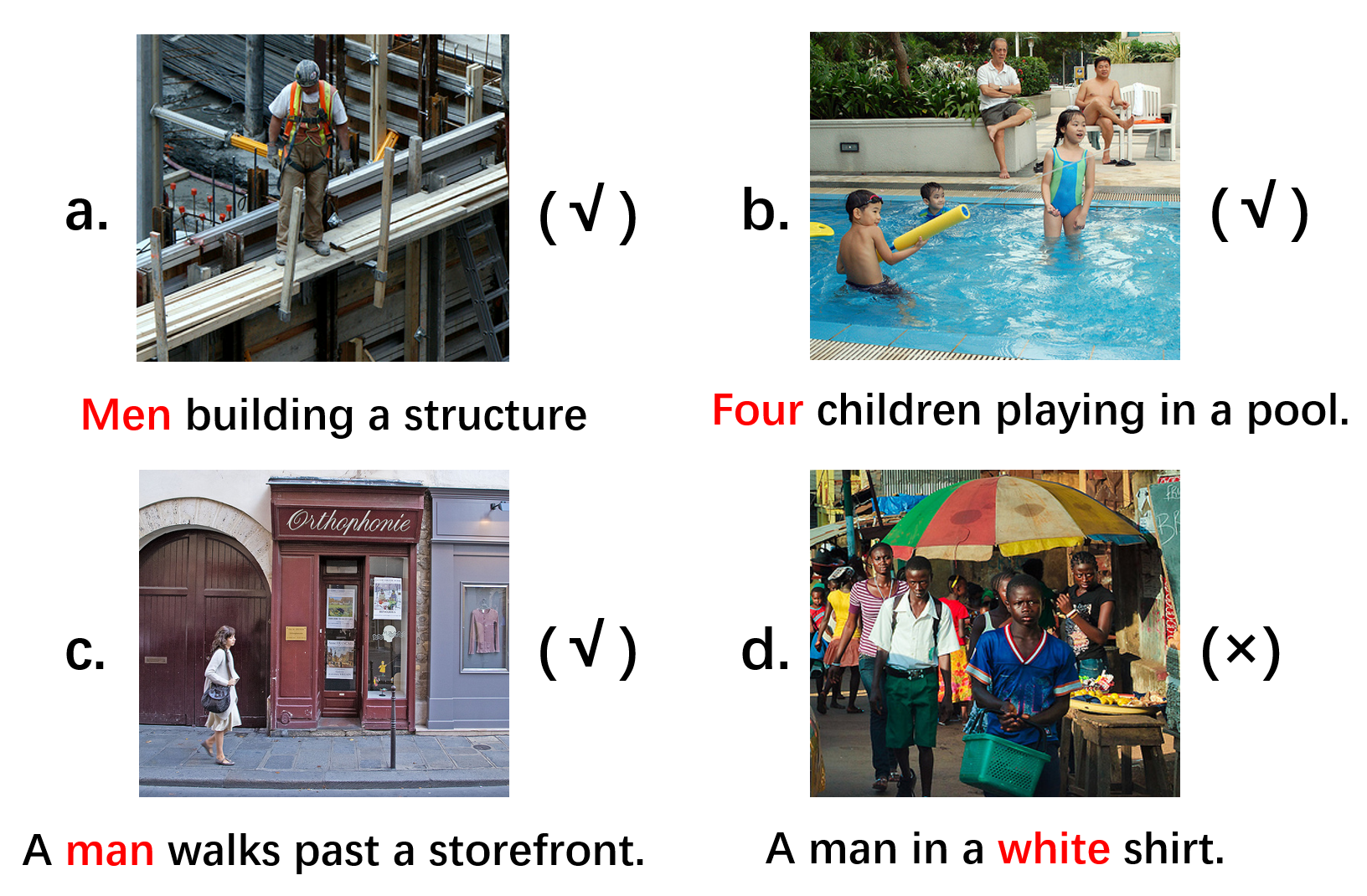} 
    \caption{Typical error cases of our multi-modal entailment model inference.
    The entailment relationship inferred by the model is remarked as the symbol "$\surd$" and the symbol "×" on the contrary.
    } 
    \label{Fig.case3} 
\end{figure}

\subsection{Case Study} \label{section:case study}

\subsubsection{Multi-Modal Entailment}

During annotating the entailment performance, we find that our multi-modal entailment model has achieved satisfactory performance in most cases. However, there is still room for improvement in a few cases.
Error cases shown in Figure~\ref{Fig.case3} represent the following typical mistakes occurred occasionally:
(\romannumeral1) Identification of the number of objects is disturbed. In regions (a) and (b), the model does not accurately measure the number of people, like `Men' and `Man';
(\romannumeral2) Wrong recognition of gender. In region (c), the person depicted in the photo is a woman;
(\romannumeral3) For scenes with multiple objects, the model may only focus on the main objects and put less attention on others. In region (d), we try to replace ``A man in a white shirt.'' with ``A woman in a green shirt.'' and find the inference result to be entailment. However, in manual annotation, we usually also focus on secondary characters and scenes;
In the future, we could use data augmentation on the text side to reduce these mistakes, thus enhancing the robustness of the proposed model.

\subsubsection{Entailment-Enhanced Retrieval}

As for the retrieval results, we find that applying entailment-enhanced training could usually make the retrieved captions more relevant and reasonable.
As shown in Figure~\ref{Fig.case4}, before applying entailment-enhanced strategy, many inappropriate descriptions exist in the retrieval results, such as ``near a fountain'' and ``concert barrier''.
Besides, vague words like ``waiting for something'' will also reduce the retrieval quality.
After training with our strategy, the number of entailed captions has increased to 3, while original positives also increased by one.
In addition, the retrieval results describe the image from multiple aspects.  
For instance, the caption ``two men are performing on a sidewalk as a crowd watches'' indicates the number of performers in the picture, while ``a man is sitting on the street playing drums on buckets'' concretely describes what is happening in the scene.

\section{Conclusion}
In this paper, we propose to apply multi-modal entailment to handle the frequent many-to-many matching problem in image-text retrieval datasets.
Our solution recognizes the relationship and utilizes the relationship in training.
Automatic and manual experiments reveal that the proposed method can consistently improve the matching performance of retrieval models.
In the future, we plan to extend our multi-modal entailment model to the video-text retrieval task.
Besides, we are devoted to handling the typical entailment errors mentioned in Section ~\ref{section:case study}.

\bibliography{aaai23}

\clearpage
\appendix

{\LARGE \textsl{Appendix:}}

\section{Examples of Entailment-Corrected Dataset} \label{example_our_dataset}

\begin{figure}[htb] 
    \centering 
    \includegraphics[width=0.5\textwidth]{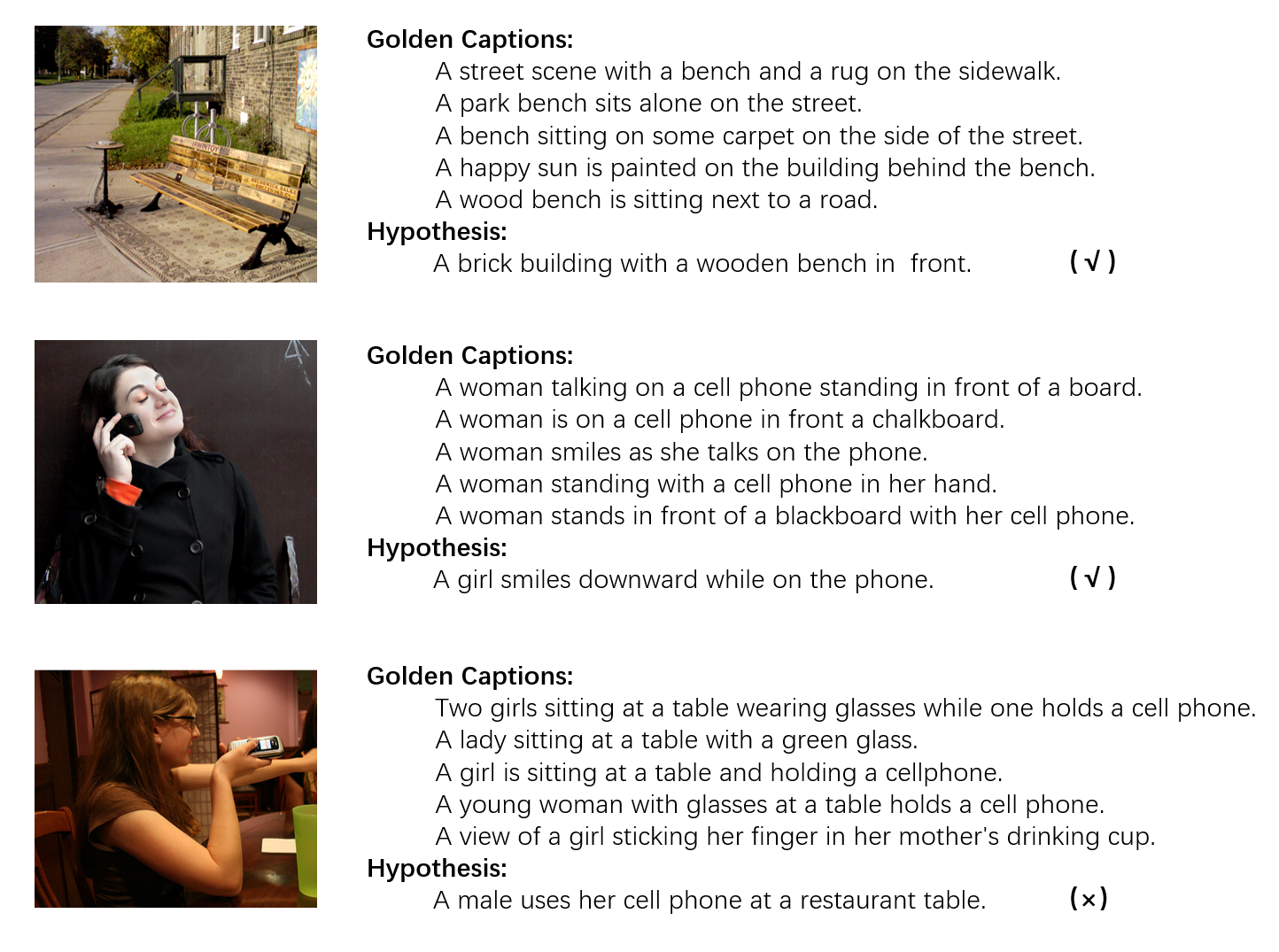} 
    \caption{Examples in our entailment-corrected dataset.
    Symbol "$\surd$" represents the entailment relationship between premise and hypothesis,
    and symbol "×" is the opposite.
    } 
    \label{Fig.examples} 
\end{figure}

Examples of our entailment-corrected dataset are shown in Figure~\ref{Fig.examples}. 
Every image corresponds to five golden captions and one hypothesis text.

\section{Datasets For Multi-modal Entailment} \label{entailment_dataset}

We constructed a training dataset for multi-modal entailment by integrating Visual entailment, Textual entailment, and Natural Language Understanding (NLU)  datasets, the components of which are shown below:

\paragraph{SNLI-VE} 
SNLI-VE is a visual entailment dataset that is constructed based on Flickr30K and SNLI.
\paragraph{CrissCrossed Caption (CxC)}
\citet{DBLP:journals/corr/abs-2004-15020} annotate the dataset  CrissCrossed Caption (CxC) based on MSCOCO to enhance the dataset of cross-modal correlations: image-image,image-text,text-text.

\paragraph{XNLI}
XNLI is a significant dataset in natural language understanding.
It contains 15 languages, and each piece of data consists of two sentences named promise and hypothesis, respectively, intending to predict the relationship between a given two sentences: entailment, contradiction, or neutral.

\paragraph{Extended COCO Validation (ECCV) }
Similar to CxC, Extended COCO Validation (ECCV) \cite{chun_eccv_2022}
is a caption dataset containing 1,261 image queries (originally 5,000) but with 17.9 positive captions per image query on average (originally 5). 
It also contains 1,332 caption queries (originally 25,000) with 8.5 positive images per caption (originally 1).

\paragraph{MRPC}

Microsoft Research Paraphrase Corpus
consists of sentence pairs automatically extracted from online news sources, with human annotations for whether the sentences in the pair are semantically equivalent \cite{dolan-brockett-2005-automatically}.
We transform the semantic similarity discriminant in sentence pairs into an entailment discriminant.

\paragraph{RTE}
Recognizing Textual Entailment is a binary entailment task similar to XNLI but with much less training data \cite{bentivogli_fifth_nodate}.

\paragraph{STS-B}
The Semantic Textual Similarity Benchmark is a collection of sentence pairs drawn from news headlines and other sources \cite{cer-etal-2017-semeval}.
They were annotated with a score from 1 to 5 denoting how similar the two sentences are in terms of semantic meaning.

\paragraph{QQP}
Quora Question Pairs is a binary classification task that aims to determine if two questions asked on Quora are semantically equivalent \cite{Chen2017QuoraQP}.

\paragraph{Text Simplification(TS)}

The text simplification task is to transform a complex sentence into a clean and clear sentence, which makes it more convenient to read and communicate \cite{kauchak-2013-improving}.
To translate the data into the form of an entailment task, we consider the existence of entailment relations between pairs of sentences in the text simplification task.

Since the labels of STS-B and CXC datasets are scores ranging from 0 to 5, we use three as a threshold and thus transform them to be usable for our task.

\section{Baseline Models For Image-Text Retrieval} \label{baseline_models}

\paragraph{ALBEF}
\cite{li_align_2021} model combines a ViT as a visual encoder and stacked 6-layer transformer blocks as text encoders.
In the image-text retrieval task,
ALBEF first aligns the unimodal image and text representation before fusing them with a multi-modal encoder.

\paragraph{CLIP}
\cite{radford_learning_2021} performs pre-training on massive noisy image-text data using a contrastive loss.
CLIP officially provides a variety of image encoders.
In our experiment,
we choose the official ViT-B/32 as our image encoder for quickly training and evaluation.

\paragraph{UNITER}
\cite{chen_uniter_2020} leverage a transformer-based architecture to learn universal representations from image and text features.
We choose UNITER-base as our pre-train model.

\end{document}